\documentclass[10pt,twocolumn,letterpaper]{article}

\usepackage{iccv}
\usepackage{times}
\usepackage{epsfig}
\usepackage{graphicx}
\usepackage{amsmath}
\usepackage{amssymb}
\usepackage{booktabs}
\usepackage[ruled,vlined]{algorithm2e}
\usepackage{pgfplots}
\usepackage{multirow}
\usepackage{subcaption}
\usepackage{threeparttable}

\usepackage{array}
\newcommand{\PreserveBackslash}[1]{\let\temp=\\#1\let\\=\temp}
\newcolumntype{C}[1]{>{\PreserveBackslash\centering}p{#1}}
\newcolumntype{R}[1]{>{\PreserveBackslash\raggedleft}p{#1}}
\newcolumntype{L}[1]{>{\PreserveBackslash\raggedright}p{#1}}

\usepackage[pagebackref=true,breaklinks=true,letterpaper=true,colorlinks,bookmarks=false]{hyperref}

\iccvfinalcopy 

\def\iccvPaperID{****} 

\ificcvfinal\fi

\begin{document}

\title{Combining Past, Present and Future: A Self-Supervised Approach for Class Incremental Learning}

\author{Xiaoshuang Chen\\
Shanghai Jiao Tong University\\
{\tt\small chenxiaoshuang@sjtu.edu.cn}
\and
Zhongyi Sun\\
Tencent Youtu Lab\\
{\tt\small zhongyisun@tencent.com}
\and
Ke Yan\\
Tencent Youtu Lab\\
{\tt\small kerwinyan@tencent.com}
\and
Shouhong Ding\\
Tencent Youtu Lab\\
{\tt\small ericshding@tencent.com}
\and
Hongtao Lu\\
Shanghai Jiao Tong University\\
{\tt\small htlu@sjtu.edu.cn}
}

\maketitle
\ificcvfinal\thispagestyle{empty}\fi

\begin{abstract}
   Class Incremental Learning (CIL) aims to handle the scenario where data of novel classes occur continuously and sequentially.
   The model should recognize the sequential novel classes while alleviating the catastrophic forgetting.
   In the self-supervised manner, it becomes more challenging to avoid the conflict between the feature embedding spaces of novel classes and old ones without any class labels.
   To address the problem, we propose a self-supervised CIL framework CPPF, meaning Combining Past, Present and Future.
   In detail, CPPF consists of a prototype clustering module (PC), an embedding space reserving module (ESR) and a multi-teacher distillation module (MTD).
   1) The PC and the ESR modules reserve embedding space for subsequent phases at the prototype level and the feature level respectively to prepare for knowledge learned in the future.
   2) The MTD module maintains the representations of the current phase without the interference of past knowledge. One of the teacher networks retains the representations of the past phases, and the other teacher network distills relation information of the current phase to the student network.
   Extensive experiments on CIFAR100 and ImageNet100 datasets demonstrate that our proposed method boosts the performance of self-supervised class incremental learning.
   We will release code in the near future.
\end{abstract}

\section{Introduction}
\label{sec:intro}

\begin{figure}[tp]
    \centering
    \includegraphics[width=1.0\linewidth]{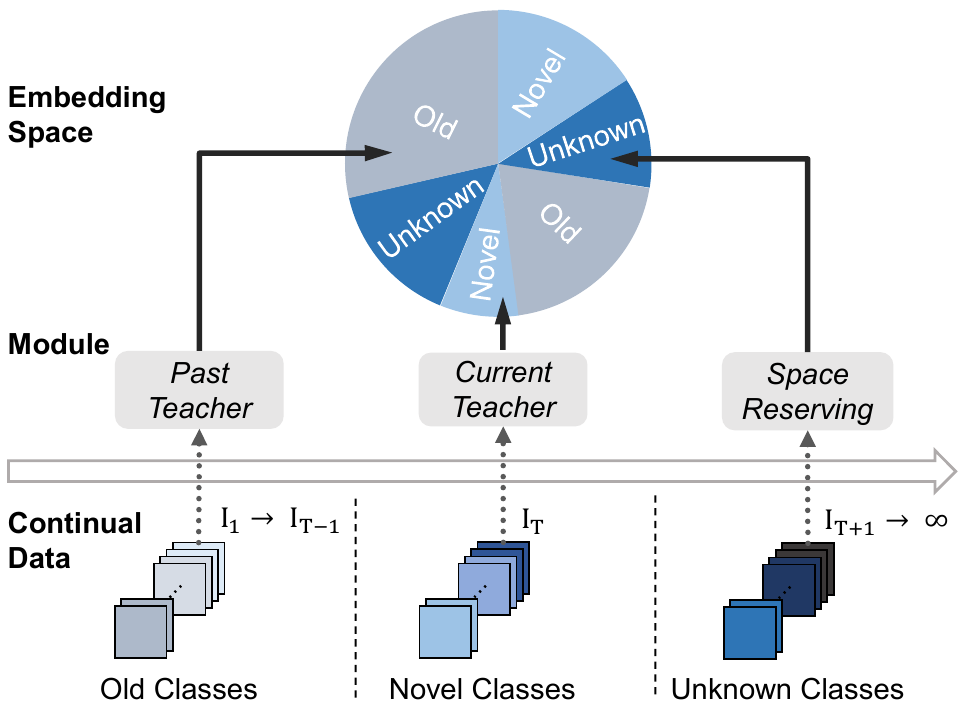}
    \caption{The motivation of our method. Our model not only preserves the past knowledge, but also handles the current knowledge properly and prepares for the future knowledge. In detail, we preserve the knowledge of old classes distilled from the past teacher, maintain the relation-level information of novel classes distilled from the current teacher, and reserve embedding space for unknown classes in subsequent phases.}
    \label{fig:moti}
\end{figure}

In the real world applications, data are usually presented in stream format, such as data from the social network.
This format requires deep neural network models to learn novel classes and obtain new knowledge incrementally.
Continual learning aims to learn from data stream sequentially, relieving catastrophic forgetting~\cite{goodfellow2013empirical}.
The catastrophic forgetting problem of neural networks implies that the knowledge learned from previous phases tends to be forgotten in the current or the future phases.
Class incremental learning (CIL) is a typical sub-domain of continual learning, where networks learn from data of novel classes while maintaining the knowledge learned previously.
Previous works~\cite{kirkpatrick2017overcoming,rebuffi2017icarl,zeng2019continual,wang2022learning} address CIL in the supervised manner.
However, the amount of stream data is often so enormous that the annotation work is laborious and time-consuming, and thus supervised CIL is not applicable to stream data in real-world unsupervised scenarios.
Therefore, exploring self-supervised CIL is a topic of research significance.
Without any class labels in the self-supervised setting, the catastrophic forgetting problem becomes more intractable, because there is no prior knowledge for the model to balance the discrimination between representations of old and new classes.

Previous works such as CaSSLe~\cite{fini2022self} address the catastrophic forgetting problem in self-supervised CIL by a distillation module.
With the distillation module, the model can preserve the knowledge learned in the past phase.
Moreover, alleviating the catastrophic forgetting problem needs not only preserving the past knowledge, but also handling the current knowledge properly and preparing for the future knowledge.
A current phase or a future phase will again become a past phase in the future.
By balancing the knowledge in the three stages, the model can avoid the catastrophic forgetting problem dynamically.
The motivation of our method is illustrated in Figure~\ref{fig:moti}.
All classes, including old classes, current novel classes and unknown classes, need to be considered together in the incremental training process.
Thus the motivation can be elaborated from three aspects.
1) The representations of old classes, considered as the past knowledge, can be preserved by the model trained in the previous phases.
2) The representations of novel classes, considered as the current knowledge, will not be interfered by the past knowledge if there is a network trained only using novel classes.
3) The representations of unknown classes, considered as the future knowledge, require reserved space to avoid conflicts with representations of known classes.
In order to combine knowledge from past, present and future, we propose a self-supervised CIL framework named CPPF.

Distilling knowledge only from the previous model as most existing works did ignores the relation among the current novel classes.
Without the conflict between the past and the current knowledge, the model can better maintain the knowledge of the current phase.
DnC~\cite{tian2021divide} trains a teacher model on each image cluster to avoid the interference with other clusters and thus learn better representations.
Similarly, a model trained using current classes alone without the past knowledge can achieve better performance on the current classes.
Thus we propose a multi-teacher distillation module (MTD), which contains two teacher networks.
One of the teacher networks distills feature-level knowledge from the previous phase to the student network, i.e., the network of the current phase, similar to CaSSLe~\cite{fini2022self}.
The other teacher network is trained only using the novel classes and maintains the relation-level information of the current classes to avoid interference with previous classes.
Combining the two teacher networks, the MTD can better handle current data relation while preserving old knowledge.

On the other hand, a good model should prepare for unknown classes which are accessible in subsequent phases.
Previous works reserve embedding space for unknown classes in supervised learning.
For instance, FACT~\cite{zhou2022forward} introduces virtual prototypes to squeeze the embedding of known classes and reserve for new ones.
Requiring class labels, these methods are unfeasible in self-supervised learning.
In this paper, we propose a prototype clustering module (PC) and an embedding space reserving module (ESR) to address this problem at both prototype-level and feature-level.
To the best of our knowledge, our work is the first to reserve embedding space for unknown classes in CIL without any class information.
Using prototypes to represent feature distribution is a common method in self-supervised learning works~\cite{caron2020unsupervised,li2020prototypical}.
The prototypes need to be adjusted dynamically when classes come incrementally.
At prototype-level, the PC introduces a clustering process for the prototypes to constrain the model to learn the prototypes incrementally.
At feature-level, the ESR divides the embedding space into two parts, one of which is reserved for subsequent phases.
A margin loss constrains the features extracted by the encoder to occupy only part of the embedding space.
No class information is used in either the PC or the ESR module.

Unlike rehearsal-based methods~\cite{gepperth2016bio,cha2021co2l,tang2021gradient,madaan2021representational} with a data memory to replay previously stored samples in the current phase, the space occupation of which is uncontrollable when the number of classes increases constantly, our method does not rely on any memory to store samples or features.

The contributions of our proposed CPPF for self-supervised CIL are as follows:

1) We propose a prototype clustering module (PC) and an embedding space reserving module (ESR) to prepare for the representations learned in the future by reserving embedding space for the next phase in CIL.

2) We propose a multi-teacher distillation module (MTD) to encourage the model to maintain both the past knowledge and the relation of the current data.

3) Extensive experiments on CIFAR100 and ImageNet100 datasets demonstrate that our method boosts the performance of self-supervised CIL.

\section{Related Work}
\label{sec:related}

\subsection{Continual Learning}

The topic of continual learning addresses the catastrophic forgetting problem of learning from data stream, to incrementally extend acquired knowledge.
Continual learning is divided into four main categories: class incremental, data incremental, task incremental and domain incremental learning.
In this paper, we focus on the class incremental learning, which implies that classes arise sequentially and only partial classes are available in each training phase.

Most current class incremental learning methods are fully or partially based on supervised learning.
Rehearsal-based methods explicitly store old samples or features while training on new classes.
GeppNet~\cite{gepperth2016bio} puts previous samples that are not easily classified into the short-term memory module.
GCR~\cite{tiwari2022gcr} proposes a novel strategy for replay buffer selection and update using a carefully designed optimization criterion.
iCaRL~\cite{rebuffi2017icarl} proposes a nearest-mean-of-exemplars classifier, a herding-based step for prioritized exemplar selection, and a representation learning step that uses the exemplars in combination with distillation.
Regularization-based methods introduce extra regularization terms for the parameter updating.
EWC~\cite{kirkpatrick2017overcoming} adds a regularization term of the difference between parameters of two phases and assigns weights according to the importance of parameters.
Sensitivity-Driven~\cite{tartaglione2018learning} quantifies the sensitivity of the parameters and gradually reduces the absolute value of the parameters with low sensitivities.
OWM~\cite{zeng2019continual} calculates the orthogonal direction of the old input space, and updates network parameters along this direction.
Other methods aim to propose new model architectures for CIL.
DGR~\cite{shin2017continual} proposes a novel framework with a cooperative dual model architecture consisting of a deep generative model and a task solving model.
DGM~\cite{ostapenko2019learning} relies on conditional generative adversarial networks with learnable connection plasticity realized with neural masking.
L2P~\cite{wang2022learning} learns a prompt pool to instruct the model conditionally and automatically selects and updates prompts from the pool in an instance-wise fashion.

Reserving embedding space for subsequent phases is an effective approach in CIL, which aims to incorporate novel classes without disrupting the discriminability of old classes.
Previous methods working on reserving embedding space require class labels for space division.
FACT~\cite{zhou2022forward} designs virtual prototypes to squeeze the embedding space of known classes and reserve for unknown classes.
The virtual prototypes act as proxies scattered among embedding space to accept possible new classes in the future.
CwD~\cite{shi2022mimicking} regularizes feature representations of each class to scatter more uniformly in the initial phase, thus mimicking the oracle model.
Nevertheless, in self-supervised manner, the lack of class labels makes embedding space reserving more challenging.
Our method divides embedding space by self-adaptive clustering, without any labels.

\subsection{Self-Supervised Class Incremental Learning}

With class labels in each phase, the model can categorize or cluster samples according to class labels.
Besides, with the number of classes in each phase, we can assign weights to different training phases.
However, in the real world, the amount of data is usually enormous and the annotation work is time-consuming.
Thus self-supervised CIL is worthy of research.

Some works partially adopt self-supervised learning.
MAS~\cite{aljundi2018memory} accumulates an importance measure for network parameters, and changes to important parameters can be penalized, effectively preventing important previous knowledge from being overwritten.
PASS~\cite{zhu2021prototype} proposes a non-exemplar based
method to overcome the catastrophic forgetting problem by memorizing and augmenting prototypes of old classes, and adopts SSL-based label augmentation.
Co2L~\cite{cha2021co2l} learns representations using the contrastive learning objective and preserves learned representations using a self-supervised distillation step.
GRCL~\cite{tang2021gradient} proposes a source discriminative constraint to improve the discriminative ability of features in the target domains and a target memorization constraint to explicitly memorize the knowledge on old target domains.

There are very few methods~\cite{madaan2021representational,fini2022self,purushwalkam2022challenges,zhuangwell,fini2022self} completely based on self-supervised learning.
LUMP~\cite{madaan2021representational} proposes Lifelong Unsupervised Mixup for unsupervised continual learning to bridge the gap between continual learning and representation learning.
MinRed~\cite{purushwalkam2022challenges} proposes minimum redundancy buffers, which maintain the least redundant samples, to learn more effective representations in self-supervised continual learning.
CaSSLe~\cite{fini2022self} proposes a distillation module to map the current state of the representations to their past state.
The module is compatible with several self-supervised models.

In general, there are already a few CIL works based on SSL, but this topic still needs more in-depth exploration.

\subsection{Distillation}

This paper is also related to knowledge distillation~\cite{hinton2015distilling}, which is a common method of distilling knowledge in a cumbersome model and compressing it into a simple model, and is widely used in many fields of deep learning.
Some CIL works use distillation to retain the knowledge learned from previous phases.
DDE~\cite{hu2021distilling} proposes to distill the causal effect, i.e., the old feature of a new sample, and shows that such distillation is causally equivalent to data replay.
Cheraghian \textit{et al.}~\cite{cheraghian2021semantic} proposes a semantically-guided knowledge distillation approach for few-shot CIL using semantic word vectors.
AFC~\cite{kang2022class} proposes a knowledge distillation algorithm for CIL with feature map weighting, which minimizes the upper bound of the loss increases over the previous tasks, which is derived by recognizing the relationship between the representation changes and the loss increases.

In addition to preserving the knowledge of the past, maintaining the relation of current data is also necessary in CIL.
There are some distillation frameworks called multi-teacher distillation~\cite{you2017learning, xiang2020learning, yuan2021reinforced}, where several expert models are trained and then distilled into a single model.
The knowledge is distilled from different teacher models and the student model can acquire information from different levels or dimensions.
Our distillation module, which can be categorized as a multi-teacher distillation module, takes both the past and the current knowledge into account.

\section{Method}
\label{sec:method}

\begin{figure*}[tp]
    \centering
    \includegraphics[width=1.0\linewidth]{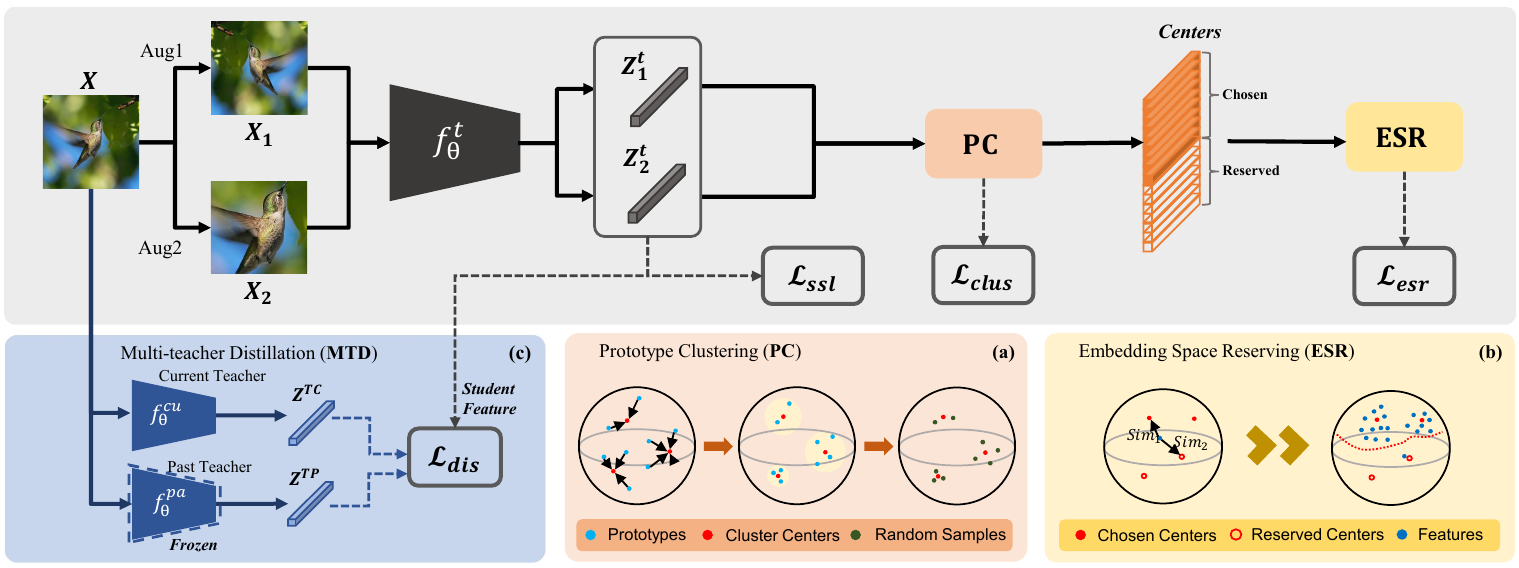}
    \caption{The overview of our method. The model consists of three modules, PC, ESR, and MTD. The network backbone can be an arbitrary model. An input sample is augmented by different data augmentation parameters to generate two views, as commonly used in self-supervised learning. The PC module generates prototypes incrementally by introducing self-adaptive cluster centers for prototypes. The ESR module reserves embedding space for unknown classes based on the distance to the cluster centers. The MTD module consists of two teacher networks. The past teacher, with frozen network parameters, maintains the representations of old classes, and the current teacher is only trained using novel classes of the current phase.}
    \label{fig:method}
\end{figure*}

In this paper, we propose a self-supervised CIL framework named CPPF, motivated by combining knowledge learned from past, present and future.
In detail, the CPPF framework consists of three components, the prototype clustering (PC) module, the embedding space reserving (ESR) module and the multi-teacher distillation (MTD) module.
In this section, we introduce the three components separately.
The overview of our method is illustrated in Figure~\ref{fig:method}.

The CPPF framework combines PC, ESR and MTD by an overall loss function, which is defined as:
\begin{equation}
    \mathcal{L} = \mathcal{L}_{ssl} + \mathcal{L}_{clus} + \alpha\mathcal{L}_{esr} + \mathcal{L}_{dis},
\end{equation}
where $\mathcal{L}_{ssl}$ is the self-supervised loss function; $\mathcal{L}_{clus}$ is the clustering loss for PC; $\mathcal{L}_{esr}$ is the margin loss for ESR; $\mathcal{L}_{dis}$ is the distillation loss for MTD.
$\alpha$ is the weight for the margin loss, and the loss weights for other losses are fixed to $1$ for simplicity.

The overall training algorithm is shown in Algorithm~\ref{alg:1}.
Our model is trained in an end-to-end manner with the overall loss.
The student network parameters $\theta$ are updated using the overall loss.
The cluster centers $c$ are updated using the clustering loss $\mathcal{L}_{clus}$ in the PC module.
The current teacher network parameters $\theta^{cu}$ are updated using the SSL loss $\mathcal{L}^{cu}_{ssl}$.
The past teacher network parameters are frozen in the training process.
In the inference stage, we only use fixed $F_{\theta}$ to extract features from test samples.

\subsection{Prototype Clustering}

Using prototypes is a common approach to represent feature distribution in self-supervised learning methods~\cite{caron2020unsupervised,li2020prototypical}.
The prototypes applied in offline learning are distributed throughout the entire embedding space.
When samples of different classes are accessed incrementally in CIL, the distribution of prototypes needs to be adjusted dynamically as the number of classes increases.

In order to generate prototypes incrementally and reserve embedding space for subsequent valid prototypes, we propose a self-adaptive prototype clustering module.
As shown in Figure~\ref{fig:method}(a), our proposed Prototype Clustering Module (PC) introduces a fixed number of cluster centers for the prototypes.
The clustering loss is the distance between each prototype and the nearest cluster center.
We use cosine similarity as the distance metric.
Suppose there are $n_c$ cluster centers and $n_p$ prototypes and then the prototype clustering loss is defined as:
\begin{equation}
    \mathcal{L}_{clus} = \frac{1}{n_p}\sum_{i=1}^{n_p}\min_{1\leq j \leq n_c}dist(f_{p_i}, f_{c_j}),
    \label{eq:PC}
\end{equation}
where $f_{p_i}$ and $f_{c_j}$ are the $i$th prototype and the $j$th cluster center; $dist(\cdot)$ is the cosine distance function.
The cluster centers and the prototypes are updated using $\mathcal{L}_{clus}$ by backpropagation with the network training. 
Although $n_c$ is fixed for training, in the clustering process, empty clusters and overlapping centers are permissible.
Thus the clustering process is self-adaptive using a fixed number of cluster centers.
Compared with other clustering algorithms with a definite number of clusters such as K-Means, which is deterministic after initialization, our clustering loss is more suitable for self-supervised learning.

Besides, storing features of previous samples in a queue is common in SSL methods~\cite{caron2020unsupervised,he2020momentum}.
The memory of the queue needs to be enlarged constantly to cope with the increase in the number of classes, otherwise the feature representation capability of the queue is weakened.
Instead of storing features, we sample previous prototypes randomly using previous cluster centers and push the sampled prototypes into a queue.
In detail, we use Gaussian random sampling, with each previous cluster center as the mean and the variance of each cluster as the variance of the Gaussian distribution.
Suppose cluster $i$ has $n_p^i$ prototypes, and then we sample $n_p^i$ random prototypes in the queue following $q_1, q_2, \dots, q_{n_p^i}\sim N(c_i,\sigma^2)$, where $\sigma^2$ is the variance of previous prototypes $p_1, p_2, \dots, p_{n_p^i}$ assigned to cluster $i$.
The queue is updated using random sampling every training epoch.
No previous samples need to be stored in our queue or extra memory.

\subsection{Embedding Space Reserving}

CPPF prepares for unknown new classes in the future through the Embedding Space Reserving Module (ESR).
This module is independent of any class information and is suitable for self-supervised learning.
As shown in Figure~\ref{fig:method}(b), the ESR uses the cluster centers in PC as anchors to divide the embedding space into two parts.
The cluster centers are randomly divided into one chosen group and the other reserved group.
The features are expected to be close to chosen centers and far from reserved centers, then the embedding space of the known classes is squeezed using the cluster centers in the reserved group as anchors for the embedding space of unknown classes.
We constrain the distance between features and different centers by a margin loss.
The chosen group is the set $C_1$ and the reserved group is the set $C_2$.
Formally, the constraint condition is $Sim_1 > \lambda Sim_2$, where $Sim_i$ is the cosine similarity between the feature and the nearest center of the group $C_i$.
The margin loss is formulated as:
\begin{equation}
\begin{split}
    & \mathcal{L}_{esr}=\frac{1}{N}\sum_{i=1}^N\mathrm{relu}(\lambda Sim_1^i-Sim_2^i),\\
    & Sim_j^i=\max_{c_k\in C_j}\frac{f_i}{|f_i|}\cdot\frac{f_{c_k}}{|f_{c_k}|}, j\in \{1, 2\},
\end{split}
\end{equation}
where $f_i$ is the feature of the $i$th sample; $N$ is the number of samples; $f_{c_j}$ is defined in the same way as Eq.~\ref{eq:PC}.
With the margin loss function, the features become close to cluster centers of $C_1$ and far from cluster centers of $C_2$.
Then the embedding space around centers of $C_2$ is reserved for subsequent classes.
The activation function ReLU truncates the gradient if $Sim_2^i >= \lambda Sim_1^i$ is satisfied, otherwise the reserved embedding space will become as large as possible.
The $\lambda$ is a sensitive hyper-parameter on balancing the proportion of the chosen embedding space and the reserved embedding space.
We carry out ablation experiments on different $\lambda$ in Sec.~\ref{subsec:res} to select the optimal $\lambda$, which is then fixed in other experiments.
The cluster centers, which are updated by $\mathcal{L}_{clus}$ in the PC module, will not be updated in the backpropagation of $\mathcal{L}_{esr}$.
With the ESR module, our method can prepare for future knowledge without forgetting the knowledge learned currently.

\begin{algorithm}[tp]
	\caption{The overall training algorithm.}
	\label{alg:algorithm1}
	\KwIn{Samples $x$ from novel classes; cluster centers $c$; previous encoder $F^{pa}_{\theta^{pa}}$ with frozen parameters $\theta^{pa}$; encoder of the current teacher network $F^{cu}_{\theta^{cu}}$ with parameters $\theta^{cu}$; loss weights $\alpha$ and $\gamma_{ep}$.}
	\KwOut{Encoder $F_{\theta}$ with parameters $\theta$.}  
	\BlankLine
	Initialize $c$, $\theta$ and $\theta^{cu}$ randomly;
	
	\For {$ep$ $\gets$ $0$ to $max\_epochs$}{
	    $f$ $\gets$ $F_{\theta}(x)$;
	    
	    $f^{pa}$ $\gets$ $F^{pa}_{\theta^{pa}}(x)$;
	    
	    $f^{cu}$ $\gets$ $F^{cu}_{\theta^{cu}}(x)$;
	    
	    
	    
	    
	    
	    
	    
	    Update $\theta$ by minimizing $\mathcal{L}_{ssl} + \mathcal{L}_{clus} + \alpha\mathcal{L}_{esr} + \mathcal{L}_{dis}$;
	    
	    Update $c$ by minimizing $\mathcal{L}_{clus}$;
	    
	    Update $\theta^{cu}$ by minimizing $\mathcal{L}_{ssl}^{cu}$;
	}
\label{alg:1}
\end{algorithm}

\subsection{Multi-teacher Distillation}

To preserve past knowledge, it is straightforward to use the previous model as the teacher network in the distillation module, but the representations of current classes tend to be interfered by past knowledge.
The best expert model for any domain needs to be trained separately~\cite{tian2021divide} with only samples from this domain.
The feature representations are free to be distributed throughout the entire embedding space without the interference of other domains.
Based on such motivation, we design a separate teacher network for the current phase and propose a multi-teacher distillation module (MTD).

In MTD, there are two teacher networks, as shown in Figure~\ref{fig:method}(c).
One of the teacher networks, called the past teacher network, is the previous model trained in the previous phase, similar to CaSSLe~\cite{fini2022self}.
Since there is no memory in our method, the samples of old classes are unavailable.
Therefore, the past teacher network can not extract credible relations among samples of novel classes.
In MTD, the past teacher network distills features instead of relations to the student network.
A predictor $g$ is used to project the representations from the embedding space of the student network to that of the past teacher network.
The distillation loss function of the past teacher network is defined as:
\begin{equation}
    \mathcal{L}_{dis}^{pa} = \frac{1}{N}\sum^N_i dist(f^{t}_i, g(f^s_i)),
\end{equation}
where $f^{t}_i$ and $f^s_i$ are features of the $i$th sample extracted from the past teacher network and the student network respectively; $dist(\cdot)$ is the cosine distance function.

The other teacher network, called the current teacher network, is trained from scratch using samples of the current phase.
Different from the past teacher network, directly distilling the features from the current teacher network to the student network is in conflict with the embedding space reserving module.
The student network needs to learn the feature relation from the current teacher network.
The current teacher network exactly maintains the relation-level information of novel classes without the interference of old knowledge.
First we compute the similarity between samples from different views, i.e., samples using different augmentations.
Then the probability distribution of feature similarity is formulated by the Softmax function:
\begin{equation}
\begin{split}
    & p_{i,k} = \frac{sim_{i, k}/\tau}{\sum_j e^{sim_{i, j}}/\tau},\\
    & sim_{i,j} = \frac{f_i^1}{|f_i^1|}\cdot\frac{f_j^2}{|f_j^2|},
\end{split}
\label{eq:softmax}
\end{equation}
where $f_i^1$ and $f_i^2$ are samples from different views; $\tau$ is the temperature in the Softmax function.
We use the Cross Entropy loss as the distillation loss of the current teacher network, which is defined as:
\begin{equation}
    \mathcal{L}_{dis}^{cu} = \frac{1}{N}\sum_i^N\sum_k(-p_{i,k}^{t} \log(p_{i,k}^s)),
\end{equation}
where $p_{i,k}^{t}$ and $p_{i,k}^s$ are the probability distributions of feature similarity in the current teacher network and the student network respectively as defined in Eq.~\ref{eq:softmax}.

The current teacher network is trained along with the student network using SSL loss.
Combining the distillation loss functions of both the past and the current teacher networks, the loss function of MTD is defined as:
\begin{equation}
    \mathcal{L}_{dis} = \mathcal{L}_{dis}^{pa} + \gamma \mathcal{L}_{dis}^{cu},
\end{equation}
where $\mathcal{L}_{ssl}^{cu}$ is the SSL loss function for the current teacher network; $\gamma$ is a dynamic weight of the loss of the current teacher network, and increases every epoch, considering that at the beginning of training the current teacher, the feature extraction capability of the network is relatively poor compared with the later training period.

\section{Experiments}
\label{sec:exp}

\subsection{Settings}
\label{subsec:set}

\noindent\textbf{Datasets:} We perform experiments on CIFAR100~\cite{krizhevsky2009learning} and ImageNet100.
CIFAR100 is a 100-class dataset with 60,000 32$\times$32 images.
ImageNet100 is a 100-class subset of ImageNet1k~\cite{deng2009imagenet}, which is a 1000-class dataset with around 1.3 million images, and the crop size of each image is 224$\times$224.

\noindent\textbf{Implementation Details:} For all our experiments, we use ResNet18~\cite{he2016deep} as the backbone network for the convenience of comparison.
We use the SGD optimizer with the warm-up and the Cosine scheduler.
The batch size is set to 256 for all the experiments.
The model is trained for 500 epochs on CIFAR100 and 400 epochs on ImageNet100.

For our MTD module, the loss weight $\gamma$ increases by the Cosine scheduler in each phase, with the initial loss weight of $0.01$ and the final loss weight of $1$.
For our ESR module, the loss weight $\alpha$ is set differently on the two datasets.
On CIFAR100 dataset, the $\alpha$ is fixed to $0.1$.
On ImageNet100 dataset, the $\alpha$ is adjusted dynamically and increases every few epochs as $0, 10^{-3}, 10^{-2}, 10^{-1}$ in each phase.

Our baseline SSL model is SwAV~\cite{caron2020unsupervised}, with the feature-based distillation module as in CaSSLe~\cite{fini2022self}.

\noindent\textbf{Evaluation Metric:} We use the average linear probing accuracy as the evaluation metric, following previous works~\cite{fini2022self,cha2021co2l}.
Formally, there are $T$ training phases in total and the linear probing accuracy of the $i$th phase after training $T$ phases is $A_T^i$, then the average linear probing accuracy $A_T$ is formulated as:
\begin{equation}
    A_T = \frac{1}{T}\sum_{i=1}^{T}A_T^i.
\end{equation}
We train a single-layer MLP for 100 epochs in the evaluation stage to compute $A_T$.

The metric Forgetting is not adopted in our experiments.
Different from supervised CIL, in self-supervised CIL, the accuracy of classes in previous phases will not decrease in subsequent phases.
Instead, the accuracy of old classes becomes higher with the increase of classes.

\begin{table}[tp]
    \centering
    \caption{The linear probing accuracy of the baseline and CPPF with different SSL methods on CIFAR100 and ImageNet100 datasets.}
    \begin{threeparttable}
    \begin{tabular}{C{1.7cm}|C{1.2cm}|c|c}
    \toprule
        SSL & Method\tnote{1} & CIFAR100 & ImageNet100 \\
        \midrule
        \multirow{2}*{SwAV~\cite{caron2020unsupervised}} & Baseline & 61.88 & 66.72 \\
        \cmidrule(){2-4}
        & CPPF & \textbf{64.11($\uparrow$2.23)} & \textbf{67.92($\uparrow$1.20)} \\
        \midrule
       Barlow & Baseline & 60.68 & 59.68 \\
       \cmidrule(){2-4}
       Twins~\cite{zbontar2021barlow} & CPPF & \textbf{60.94($\uparrow$0.26)} & \textbf{63.6($\uparrow$3.92)} \\
        \midrule
        \multirow{2}*{SimCLR~\cite{chen2020simple}} & Baseline & 59.16 & 62.22 \\
        \cmidrule(){2-4}
        & CPPF & \textbf{60.96($\uparrow$1.80)} & \textbf{66.04($\uparrow$3.82)} \\
    \bottomrule    
    \end{tabular}
    \end{threeparttable}
    \begin{tablenotes}
    \small
    \item{1} The results of the baseline are based on our implementation.
    \end{tablenotes}
    \label{tab:loss}
\end{table}

\begin{table}[tp]
    \centering
    \caption{Comparison with state-of-the-art methods on 5-phase experiments. The evaluation metric is the linear probing accuracy.}
    \begin{threeparttable}
    \begin{tabular}{C{5cm}|c}
    \toprule
        Methods\tnote{1} & CIFAR100\\
        \midrule
        CaSSLe (Barlow Twins~\cite{zbontar2021barlow}) &  60.4 \\
        CaSSLe (SwAV~\cite{caron2020unsupervised}) & 57.8 \\
        CaSSLe (BYOL~\cite{grill2020bootstrap}) & 62.2 \\
        CaSSLe (VICReg~\cite{bardes2021vicreg}) & 53.6 \\
        CaSSLe (MoCoV2+~\cite{he2020momentum,chen2020improved}) & 59.5 \\
        CaSSLe (SimCLR~\cite{chen2020simple}) & 58.3 \\
        CPPF (SwAV~\cite{caron2020unsupervised}) & \textbf{64.1} \\
    \bottomrule
    \end{tabular}
    \end{threeparttable}
    \begin{tablenotes}
    \small
    \item{1} The results of CaSSLe are from the original paper.
    \end{tablenotes}
    \label{tab:sota}
\end{table}

\begin{table}[tp]
    \centering
    \caption{Comparison of our method and the baseline on 5-phase and 10-phase experiments. The evaluation metric is the linear probing accuracy.}
    \begin{threeparttable}
    \begin{tabular}{c|c|c|c}
    \toprule
        Method\tnote{1} & Phases & CIFAR100 & ImageNet100\\
        \midrule
        Baseline & 5 & 61.88 & 66.72\\
        CPPF & 5 & \textbf{64.11($\uparrow$2.23)} & \textbf{67.92($\uparrow$1.20)}\\
        \midrule
        Baseline & 10 & 58.66 & 61.66 \\
        CPPF & 10 & \textbf{59.92($\uparrow$1.26)} & \textbf{62.18($\uparrow$0.52)} \\
    \bottomrule
    \end{tabular}
    \end{threeparttable}
    \begin{tablenotes}
    \small
    \item{1} The results of the baseline are based on our implementation.
    \end{tablenotes}
    \label{tab:10task}
\end{table}

\begin{figure*}[tp]
\centering

\begin{subfigure}[t]{0.4\linewidth}
\begin{tikzpicture}[scale=1]
    \tikzstyle{every node}=[font=\footnotesize, scale=0.9]
    \begin{axis}[
        x = 1cm,
        y = 0.245cm,
        xlabel=Phase,
        ylabel=Top1 Acc.,
        xtick = {1,2,3,4,5},
        tick align=outside,
        legend style={at={(0.8,0.25)},anchor=north},
        font=\footnotesize
        ]
    \draw[color=gray, opacity=0.5](-100,97)--(500,97);
    \draw[color=gray, opacity=0.5](-100,147)--(500,147);
    \draw[color=gray, opacity=0.5](-100,47)--(500,47);
    \draw[color=gray, opacity=0.5](-100,-3)--(500,-3);
    \draw[color=gray, opacity=0.5](0,-100)--(0,800);
    \draw[color=gray, opacity=0.5](100,-100)--(100,800);
    \draw[color=gray, opacity=0.5](200,-100)--(200,800);
    \draw[color=gray, opacity=0.5](300,-100)--(300,800);
    \draw[color=gray, opacity=0.5](400,-100)--(400,800);
    \addplot[,thick,mark=*,red] plot coordinates { 
        (1,50.29)
        (2,57.72)
        (3,59.99)
        (4,61.38)
        (5,61.88)
    };
    \addlegendentry{Baseline}
    
    \addplot[,thick,mark=triangle,blue] plot coordinates { 
        (1,50.86)
        (2,58.54)
        (3,61.73)
        (4,63.18)
        (5,64.11)
    };
    \addlegendentry{CPPF}
    
    \end{axis}
\end{tikzpicture}
\caption{Top1 accuracy on the 5-phase CIFAR100.}
\label{fig:sub1}
\end{subfigure}	
\begin{subfigure}[t]{0.4\linewidth}
\begin{tikzpicture}[scale=1]
    \tikzstyle{every node}=[font=\footnotesize, scale=0.9]
    \begin{axis}[
        x = 0.5cm,
        y = 0.137cm,
        xlabel=Phase,
        ylabel=Top1 Acc.,
        xtick = {1,2,3,4,5,6,7,8,9,10},
        tick align=outside,
        legend style={at={(0.8,0.25)},anchor=north},
        font=\footnotesize
        ]
    \draw[color=gray, opacity=0.5](-100,247)--(1000,247);
    \draw[color=gray, opacity=0.5](-100,147)--(1000,147);
    \draw[color=gray, opacity=0.5](-100,47)--(1000,47);
    \draw[color=gray, opacity=0.5](0,-100)--(0,800);
    \draw[color=gray, opacity=0.5](200,-100)--(200,800);
    \draw[color=gray, opacity=0.5](400,-100)--(400,800);
    \draw[color=gray, opacity=0.5](600,-100)--(600,800);
    \draw[color=gray, opacity=0.5](800,-100)--(800,800);
    \addplot[,thick,mark=*,red] plot coordinates { 
        (1,35.67)
        (2,46.63)
        (3,52.28)
        (4,54.59)
        (5,56.83)
        (6,57.79)
        (7,57.93)
        (8,58.42)
        (9,58.26)
        (10,58.66)
    };
    \addlegendentry{Baseline}
    
    \addplot[,thick,mark=triangle,blue] plot coordinates { 
        (1,35.27)
        (2,48.87)
        (3,54.49)
        (4,56.33)
        (5,58.3)
        (6,58.65)
        (7,58.88)
        (8,59.66)
        (9,59.65)
        (10,59.92)
    };
    \addlegendentry{CPPF}
    
    \end{axis}
\end{tikzpicture}
\caption{Top1 accuracy on the 10-phase CIFAR100.}
\label{fig:sub2}
\end{subfigure}
\caption{The results of the incremental linear evaluation. We compare our method and the baseline on the average accuracy of all classes. We evaluate the accuracy by linear probing after each phase on (a) 5-phase and (b) 10-phase CIFAR100.}
\label{fig:linear}
\end{figure*}
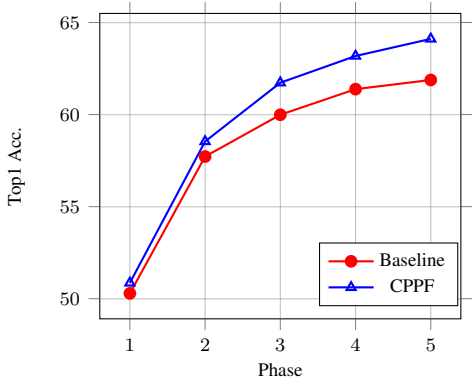
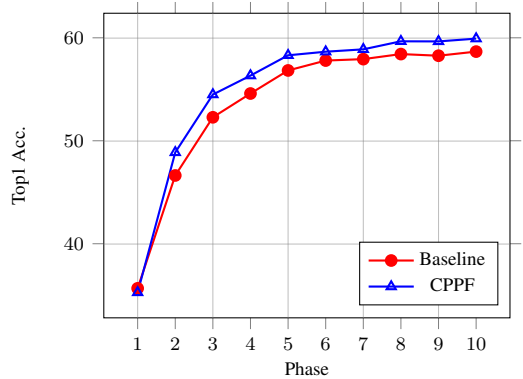

\subsection{Results}
\label{subsec:res}

\noindent\textbf{Benchmark Comparison:} Our method can be easily transplanted to different SSL methods and loss functions.
We test the effects of our method on frequently-used SSL loss functions, including the loss in Barlow Twins~\cite{zbontar2021barlow} and SimCLR~\cite{chen2020simple}.
Table~\ref{tab:loss} shows that our method is robust to the variation of SSL loss functions on both datasets.
The results demonstrate the universality of our method.

We compare the performance of our method and state-of-the-art methods on 5-phase experiments in Table~\ref{tab:sota}.
Very few methods address CIL completely based on self-supervised learning, and to the best of our knowledge, only CaSSLe~\cite{fini2022self} has the same experiment settings as ours.
We compare the result of our method with results on different SSL models in CaSSLe~\cite{fini2022self}.
It is observed that our method achieves state-of-the-art performance on 5-phase CIFAR100 dataset.

\noindent\textbf{Robustness:} We compare the performance of our proposed method with the baseline model on CIFAR100 dataset and ImageNet100 dataset.
The results of the 5-phase experiments are shown in Table~\ref{tab:10task}.
It is observed that our proposed method improves the performance of self-supervised CIL on the two datasets.

The incremental sequence length of stream data varies in practical application scenarios, especially in a self-supervised manner.
The total number of classes and the number of classes in each phase are unavailable in self-supervised learning.
This fact requires our CIL model to cope with different numbers of phases.
Therefore, we also carry out experiments with a longer sequence.
The results of 10-phase experiments on CIFAR100 and ImageNet100 are shown in Table~\ref{tab:10task}.
It is demonstrated that our method is applicable to different sequence lengths in CIL.

As illustrated in Figure~\ref{fig:linear}, our method affects the linear probing accuracy at each phase.
Our method achieves higher overall accuracy in all classes in every phase compared with the baseline.
This improvement proves that our method reaches a better balance among knowledge learned from past, present and future.

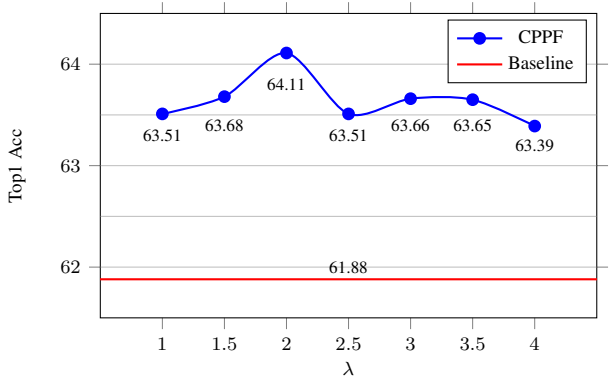
\begin{figure}[tp]
\centering
\begin{tikzpicture}[scale=1]
\tikzstyle{every node}=[font=\footnotesize, scale=0.9]
\begin{axis}[
    y = 1.35cm,
    x = 1.65cm,
    xlabel=$\lambda$,
    ylabel=Top1 Acc,
    xtick = {1,1.5,2,2.5,3,3.5,4},
    ymax=64.5,
    ymin=61.5,
    xmax=4.5,
    xmin=0.5,
    tick align=outside,
    font=\footnotesize,
    ]
    \draw[color=gray, opacity=0.5](-100,250)--(500,250);
    \draw[color=gray, opacity=0.5](-100,200)--(500,200);
    \draw[color=gray, opacity=0.5](-100,150)--(500,150);
    \draw[color=gray, opacity=0.5](-100,100)--(500,100);
    \draw[color=gray, opacity=0.5](-100,50)--(500,50);

\addplot[smooth,thick,mark=*,blue] plot coordinates { 
    (1,63.51)
    (1.5,63.68)
    (2,64.11)
    (2.5,63.51)
    (3,63.66)
    (3.5,63.65)
    (4,63.39)
};

\addplot[thick,red] plot coordinates { 
    (0,61.88)
    (5,61.88)
};

\node [minimum width=3cm,minimum height=1.5cm, font=\footnotesize, scale=0.9](anchor)at(100,180){63.51};
\node [minimum width=3cm,minimum height=1.5cm, font=\footnotesize, scale=0.9](anchor)at(150,190){63.68};
\node [minimum width=3cm,minimum height=1.5cm, font=\footnotesize, scale=0.9](anchor)at(200,230){64.11};
\node [minimum width=3cm,minimum height=1.5cm, font=\footnotesize, scale=0.9](anchor)at(250,180){63.51};
\node [minimum width=3cm,minimum height=1.5cm, font=\footnotesize, scale=0.9](anchor)at(300,190){63.66};
\node [minimum width=3cm,minimum height=1.5cm, font=\footnotesize, scale=0.9](anchor)at(350,190){63.65};
\node [minimum width=3cm,minimum height=1.5cm, font=\footnotesize, scale=0.9](anchor)at(400,170){63.39};

\node [minimum width=3cm,minimum height=1.5cm, font=\footnotesize, scale=0.9](anchor)at(250,50){61.88};

\legend{CPPF,Baseline}
\end{axis}
\end{tikzpicture}
\caption{The results with different $\lambda$ on the 5-phase CIFAR100 dataset.}
\label{fig:lambda}
\end{figure}

\begin{figure}[tp]
\centering
\begin{tikzpicture}[scale=1]
\tikzstyle{every node}=[font=\footnotesize, scale=0.9]
\begin{axis}[
    y = 1.35cm,
    x = 0.165cm,
    xlabel=The proportion of $C_1$ (\%),
    ylabel=Top1 Acc,
    xtick = {40,45,50,55,60,65,70},
    ymax=64.5,
    ymin=61.5,
    xmax=75,
    xmin=35,
    tick align=outside,
    font=\footnotesize
    ]
    \draw[color=gray, opacity=0.5](-100,250)--(500,250);
    \draw[color=gray, opacity=0.5](-100,200)--(500,200);
    \draw[color=gray, opacity=0.5](-100,150)--(500,150);
    \draw[color=gray, opacity=0.5](-100,100)--(500,100);
    \draw[color=gray, opacity=0.5](-100,50)--(500,50);

\addplot[smooth,thick,mark=*,blue] plot coordinates { 
    (40,63.77)
    (45,63.63)
    (50,64.11)
    (55,63.89)
    (60,63.68)
    (65,63.5)
    (70,63.36)
};

\addplot[thick,red] plot coordinates { 
    (30,61.88)
    (80,61.88)
};
\node [minimum width=3cm,minimum height=1.5cm, font=\footnotesize, scale=0.9](anchor)at(100,200){63.77};
\node [minimum width=3cm,minimum height=1.5cm, font=\footnotesize, scale=0.9](anchor)at(150,190){63.63};
\node [minimum width=3cm,minimum height=1.5cm, font=\footnotesize, scale=0.9](anchor)at(200,230){64.11};
\node [minimum width=3cm,minimum height=1.5cm, font=\footnotesize, scale=0.9](anchor)at(250,210){63.89};
\node [minimum width=3cm,minimum height=1.5cm, font=\footnotesize, scale=0.9](anchor)at(300,190){63.68};
\node [minimum width=3cm,minimum height=1.5cm, font=\footnotesize, scale=0.9](anchor)at(350,175){63.5};
\node [minimum width=3cm,minimum height=1.5cm, font=\footnotesize, scale=0.9](anchor)at(400,160){63.36};

\node [minimum width=3cm,minimum height=1.5cm, font=\footnotesize, scale=0.9](anchor)at(250,50){61.88};

\legend{CPPF,Baseline}
\end{axis}
\end{tikzpicture}
\caption{The results with different proportions of the chosen  cluster centers $C_1$ on the 5-phase CIFAR100 dataset.}
\label{fig:proportion}
\end{figure}
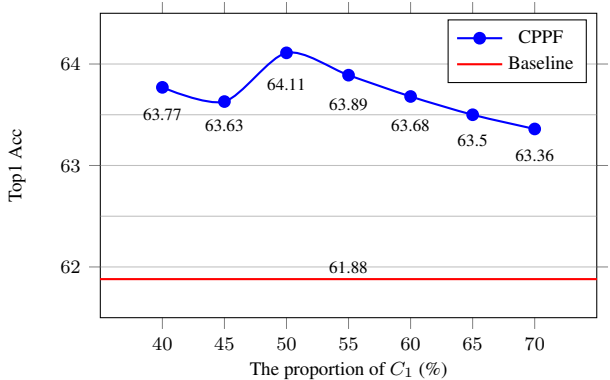

\begin{table*}[tp]
    \centering
    \caption{The ablation study of the 5-phase average linear probing accuracy on CIFAR100 and ImageNet100 datasets.}
    \begin{tabular}{c|C{0.8cm}|C{0.8cm}|C{0.8cm}|c|c}
    \toprule
        Method & PC & ESR & MTD & CIFAR100 & ImageNet100 \\
        \midrule
        Baseline&& & & 61.88 & 66.72\\
        Baseline+PC&$\surd$ & & & 63.29 ($\uparrow$1.41) & 67.30 ($\uparrow$0.58)\\
        Baseline+PC+ESR&$\surd$ & $\surd$ & & 63.69 ($\uparrow$1.81) & 67.76 ($\uparrow$1.04)\\
        Baseline+PC+MTD&$\surd$ & & $\surd$ & 63.95 ($\uparrow$2.07) & 67.44 ($\uparrow$0.72) \\
        CPPF&$\surd$ & $\surd$ & $\surd$ & \textbf{64.11 ($\uparrow$2.23)} & \textbf{67.92 ($\uparrow$1.20)} \\
    \bottomrule
    \end{tabular}
    \label{tab:ablation}
\end{table*}

\noindent\textbf{Ablation Study:} There are three modules in our proposed method, and we conduct experiments to analyze the effectiveness of each component.
The experiments are based on CIFAR100 and ImageNet100 datasets, with 5 phases for each dataset.
Experiments on the same dataset are implemented with the same settings.
The results are shown in Table~\ref{tab:ablation}.
It is observed that all three modules can improve the performance of CIL, and they are compatible with each other.
This is attributed to the motivation of the three modules.
The PC and the ESR modules aim to prepare for future knowledge by squeezing the embedding space of known classes and reserving space for new classes in the future.
The past teacher in MTD aims to preserve the past knowledge learned in previous phases, and the current teacher in MTD aims to maintain the relation of representations of novel classes in the current phase.
In consequence, by combining the three modules, our method reaches a better balance between knowledge of different phases.

In the margin loss of our ESR module, the hyper-parameter $\lambda$ decides the proportion of the reserved space for unknown classes.
We evaluate the impact of different ratios $\lambda$ on CIFAR100 dataset and show the results in Figure~\ref{fig:lambda}.
As expected, the hyper-parameter $\lambda$ affects the performance of our method, because $\lambda$ can balance the proportion of the chosen space and the reserved space.
The optimal $\lambda$ demonstrates that restricted embedding space of certain phases leads to congested representations distributions and poor discrimination of certain classes.
The $\lambda$ is set to $2$ in our experiments.
It is also observed that CPPF can achieve much better performance with different settings of $\lambda$ compared with the baseline, which proves the robustness of CPPF to hyper-parameters.

Besides, the proportion of the chosen cluster centers among all cluster centers also balances the chosen space for the current phase and the reserved space for subsequent phases.
We carry out experiments to test the influence of different proportions and illustrate the results in Figure~\ref{fig:proportion}.
The optimal proportion proves that excessive cluster centers in $C_1$ or $C_2$ lead to lower accuracy due to impairing the balance of learning past and current knowledge.
Nevertheless, our method still achieves significant improvement even with sub-optimal settings.
We simply set the proportion to $50\%$ for all the experiments for easier comparison.



\section{Conclusion}

To overcome the common catastrophic forget problem in self-supervised CIL, we propose a framework named CPPF.
CPPF can preserve past knowledge, maintain current knowledge without interference, and prepare for future knowledge.
Comprehensive experiments demonstrate that our method improves the performance of self-supervised class incremental learning.

There is still room for improvement in the implementation details of the embedding space reserving module.
For instance, we can design an algorithm for selecting the chosen centers and the reserved centers instead of random selection.
Besides, we can design a new margin loss that has no data-sensitive hyper-parameters.
We leave these possible improvements as future works.

{\small
\bibliographystyle{ieee_fullname}
\bibliography{egbib}

\begin{thebibliography}{10}\itemsep=-1pt

\bibitem{aljundi2018memory}
Rahaf Aljundi, Francesca Babiloni, Mohamed Elhoseiny, Marcus Rohrbach, and
  Tinne Tuytelaars.
\newblock Memory aware synapses: Learning what (not) to forget.
\newblock In {\em Proceedings of the European Conference on Computer Vision
  (ECCV)}, pages 139--154, 2018.

\bibitem{bardes2021vicreg}
Adrien Bardes, Jean Ponce, and Yann LeCun.
\newblock Vicreg: Variance-invariance-covariance regularization for
  self-supervised learning.
\newblock {\em arXiv preprint arXiv:2105.04906}, 2021.

\bibitem{caron2020unsupervised}
Mathilde Caron, Ishan Misra, Julien Mairal, Priya Goyal, Piotr Bojanowski, and
  Armand Joulin.
\newblock Unsupervised learning of visual features by contrasting cluster
  assignments.
\newblock {\em Advances in Neural Information Processing Systems},
  33:9912--9924, 2020.

\bibitem{cha2021co2l}
Hyuntak Cha, Jaeho Lee, and Jinwoo Shin.
\newblock Co2l: Contrastive continual learning.
\newblock In {\em Proceedings of the IEEE/CVF International Conference on
  Computer Vision}, pages 9516--9525, 2021.

\bibitem{chen2020simple}
Ting Chen, Simon Kornblith, Mohammad Norouzi, and Geoffrey Hinton.
\newblock A simple framework for contrastive learning of visual
  representations.
\newblock In {\em International conference on machine learning}, pages
  1597--1607. PMLR, 2020.

\bibitem{chen2020improved}
Xinlei Chen, Haoqi Fan, Ross Girshick, and Kaiming He.
\newblock Improved baselines with momentum contrastive learning.
\newblock {\em arXiv preprint arXiv:2003.04297}, 2020.

\bibitem{cheraghian2021semantic}
Ali Cheraghian, Shafin Rahman, Pengfei Fang, Soumava~Kumar Roy, Lars Petersson,
  and Mehrtash Harandi.
\newblock Semantic-aware knowledge distillation for few-shot class-incremental
  learning.
\newblock In {\em Proceedings of the IEEE/CVF Conference on Computer Vision and
  Pattern Recognition}, pages 2534--2543, 2021.

\bibitem{deng2009imagenet}
Jia Deng, Wei Dong, Richard Socher, Li-Jia Li, Kai Li, and Li Fei-Fei.
\newblock Imagenet: A large-scale hierarchical image database.
\newblock In {\em 2009 IEEE conference on computer vision and pattern
  recognition}, pages 248--255. Ieee, 2009.

\bibitem{fini2022self}
Enrico Fini, Victor G~Turrisi da Costa, Xavier Alameda-Pineda, Elisa Ricci,
  Karteek Alahari, and Julien Mairal.
\newblock Self-supervised models are continual learners.
\newblock In {\em Proceedings of the IEEE/CVF Conference on Computer Vision and
  Pattern Recognition}, pages 9621--9630, 2022.

\bibitem{gepperth2016bio}
Alexander Gepperth and Cem Karaoguz.
\newblock A bio-inspired incremental learning architecture for applied
  perceptual problems.
\newblock {\em Cognitive Computation}, 8(5):924--934, 2016.

\bibitem{goodfellow2013empirical}
Ian~J Goodfellow, Mehdi Mirza, Da Xiao, Aaron Courville, and Yoshua Bengio.
\newblock An empirical investigation of catastrophic forgetting in
  gradient-based neural networks.
\newblock {\em arXiv preprint arXiv:1312.6211}, 2013.

\bibitem{grill2020bootstrap}
Jean-Bastien Grill, Florian Strub, Florent Altch{\'e}, Corentin Tallec, Pierre
  Richemond, Elena Buchatskaya, Carl Doersch, Bernardo Avila~Pires, Zhaohan
  Guo, Mohammad Gheshlaghi~Azar, et~al.
\newblock Bootstrap your own latent-a new approach to self-supervised learning.
\newblock {\em Advances in neural information processing systems},
  33:21271--21284, 2020.

\bibitem{he2020momentum}
Kaiming He, Haoqi Fan, Yuxin Wu, Saining Xie, and Ross Girshick.
\newblock Momentum contrast for unsupervised visual representation learning.
\newblock In {\em Proceedings of the IEEE/CVF conference on computer vision and
  pattern recognition}, pages 9729--9738, 2020.

\bibitem{he2016deep}
Kaiming He, Xiangyu Zhang, Shaoqing Ren, and Jian Sun.
\newblock Deep residual learning for image recognition.
\newblock In {\em Proceedings of the IEEE conference on computer vision and
  pattern recognition}, pages 770--778, 2016.

\bibitem{hinton2015distilling}
Geoffrey Hinton, Oriol Vinyals, Jeff Dean, et~al.
\newblock Distilling the knowledge in a neural network.
\newblock {\em arXiv preprint arXiv:1503.02531}, 2(7), 2015.

\bibitem{hu2021distilling}
Xinting Hu, Kaihua Tang, Chunyan Miao, Xian-Sheng Hua, and Hanwang Zhang.
\newblock Distilling causal effect of data in class-incremental learning.
\newblock In {\em Proceedings of the IEEE/CVF Conference on Computer Vision and
  Pattern Recognition}, pages 3957--3966, 2021.

\bibitem{kang2022class}
Minsoo Kang, Jaeyoo Park, and Bohyung Han.
\newblock Class-incremental learning by knowledge distillation with adaptive
  feature consolidation.
\newblock In {\em Proceedings of the IEEE/CVF Conference on Computer Vision and
  Pattern Recognition}, pages 16071--16080, 2022.

\bibitem{kirkpatrick2017overcoming}
James Kirkpatrick, Razvan Pascanu, Neil Rabinowitz, Joel Veness, Guillaume
  Desjardins, Andrei~A Rusu, Kieran Milan, John Quan, Tiago Ramalho, Agnieszka
  Grabska-Barwinska, et~al.
\newblock Overcoming catastrophic forgetting in neural networks.
\newblock {\em Proceedings of the national academy of sciences},
  114(13):3521--3526, 2017.

\bibitem{krizhevsky2009learning}
Alex Krizhevsky, Geoffrey Hinton, et~al.
\newblock Learning multiple layers of features from tiny images.
\newblock 2009.

\bibitem{li2020prototypical}
Junnan Li, Pan Zhou, Caiming Xiong, and Steven~CH Hoi.
\newblock Prototypical contrastive learning of unsupervised representations.
\newblock {\em arXiv preprint arXiv:2005.04966}, 2020.

\bibitem{madaan2021representational}
Divyam Madaan, Jaehong Yoon, Yuanchun Li, Yunxin Liu, and Sung~Ju Hwang.
\newblock Representational continuity for unsupervised continual learning.
\newblock In {\em International Conference on Learning Representations}, 2021.

\bibitem{ostapenko2019learning}
Oleksiy Ostapenko, Mihai Puscas, Tassilo Klein, Patrick Jahnichen, and Moin
  Nabi.
\newblock Learning to remember: A synaptic plasticity driven framework for
  continual learning.
\newblock In {\em Proceedings of the IEEE/CVF conference on computer vision and
  pattern recognition}, pages 11321--11329, 2019.

\bibitem{purushwalkam2022challenges}
Senthil Purushwalkam, Pedro Morgado, and Abhinav Gupta.
\newblock The challenges of continuous self-supervised learning.
\newblock {\em arXiv preprint arXiv:2203.12710}, 2022.

\bibitem{rebuffi2017icarl}
Sylvestre-Alvise Rebuffi, Alexander Kolesnikov, Georg Sperl, and Christoph~H
  Lampert.
\newblock icarl: Incremental classifier and representation learning.
\newblock In {\em Proceedings of the IEEE conference on Computer Vision and
  Pattern Recognition}, pages 2001--2010, 2017.

\bibitem{shi2022mimicking}
Yujun Shi, Kuangqi Zhou, Jian Liang, Zihang Jiang, Jiashi Feng, Philip~HS Torr,
  Song Bai, and Vincent~YF Tan.
\newblock Mimicking the oracle: An initial phase decorrelation approach for
  class incremental learning.
\newblock In {\em Proceedings of the IEEE/CVF Conference on Computer Vision and
  Pattern Recognition}, pages 16722--16731, 2022.

\bibitem{shin2017continual}
Hanul Shin, Jung~Kwon Lee, Jaehong Kim, and Jiwon Kim.
\newblock Continual learning with deep generative replay.
\newblock {\em Advances in neural information processing systems}, 30, 2017.

\bibitem{tang2021gradient}
Shixiang Tang, Peng Su, Dapeng Chen, and Wanli Ouyang.
\newblock Gradient regularized contrastive learning for continual domain
  adaptation.
\newblock In {\em Proceedings of the AAAI Conference on Artificial
  Intelligence}, volume~35, pages 2665--2673, 2021.

\bibitem{tartaglione2018learning}
Enzo Tartaglione, Skjalg Leps{\o}y, Attilio Fiandrotti, and Gianluca Francini.
\newblock Learning sparse neural networks via sensitivity-driven
  regularization.
\newblock {\em Advances in neural information processing systems}, 31, 2018.

\bibitem{tian2021divide}
Yonglong Tian, Olivier~J Henaff, and A{\"a}ron van~den Oord.
\newblock Divide and contrast: Self-supervised learning from uncurated data.
\newblock In {\em Proceedings of the IEEE/CVF International Conference on
  Computer Vision}, pages 10063--10074, 2021.

\bibitem{tiwari2022gcr}
Rishabh Tiwari, Krishnateja Killamsetty, Rishabh Iyer, and Pradeep Shenoy.
\newblock Gcr: Gradient coreset based replay buffer selection for continual
  learning.
\newblock In {\em Proceedings of the IEEE/CVF Conference on Computer Vision and
  Pattern Recognition}, pages 99--108, 2022.

\bibitem{wang2022learning}
Zifeng Wang, Zizhao Zhang, Chen-Yu Lee, Han Zhang, Ruoxi Sun, Xiaoqi Ren,
  Guolong Su, Vincent Perot, Jennifer Dy, and Tomas Pfister.
\newblock Learning to prompt for continual learning.
\newblock In {\em Proceedings of the IEEE/CVF Conference on Computer Vision and
  Pattern Recognition}, pages 139--149, 2022.

\bibitem{xiang2020learning}
Liuyu Xiang, Guiguang Ding, and Jungong Han.
\newblock Learning from multiple experts: Self-paced knowledge distillation for
  long-tailed classification.
\newblock In {\em European Conference on Computer Vision}, pages 247--263.
  Springer, 2020.

\bibitem{you2017learning}
Shan You, Chang Xu, Chao Xu, and Dacheng Tao.
\newblock Learning from multiple teacher networks.
\newblock In {\em Proceedings of the 23rd ACM SIGKDD International Conference
  on Knowledge Discovery and Data Mining}, pages 1285--1294, 2017.

\bibitem{yuan2021reinforced}
Fei Yuan, Linjun Shou, Jian Pei, Wutao Lin, Ming Gong, Yan Fu, and Daxin Jiang.
\newblock Reinforced multi-teacher selection for knowledge distillation.
\newblock In {\em Proceedings of the AAAI Conference on Artificial
  Intelligence}, volume~35, pages 14284--14291, 2021.

\bibitem{zbontar2021barlow}
Jure Zbontar, Li Jing, Ishan Misra, Yann LeCun, and St{\'e}phane Deny.
\newblock Barlow twins: Self-supervised learning via redundancy reduction.
\newblock In {\em International Conference on Machine Learning}, pages
  12310--12320. PMLR, 2021.

\bibitem{zeng2019continual}
Guanxiong Zeng, Yang Chen, Bo Cui, and Shan Yu.
\newblock Continual learning of context-dependent processing in neural
  networks.
\newblock {\em Nature Machine Intelligence}, 1(8):364--372, 2019.

\bibitem{zhou2022forward}
Da-Wei Zhou, Fu-Yun Wang, Han-Jia Ye, Liang Ma, Shiliang Pu, and De-Chuan Zhan.
\newblock Forward compatible few-shot class-incremental learning.
\newblock In {\em Proceedings of the IEEE/CVF Conference on Computer Vision and
  Pattern Recognition}, pages 9046--9056, 2022.

\bibitem{zhu2021prototype}
Fei Zhu, Xu-Yao Zhang, Chuang Wang, Fei Yin, and Cheng-Lin Liu.
\newblock Prototype augmentation and self-supervision for incremental learning.
\newblock In {\em Proceedings of the IEEE/CVF Conference on Computer Vision and
  Pattern Recognition}, pages 5871--5880, 2021.

\bibitem{zhuangwell}
Chengxu Zhuang, Violet Xiang, Yoon Bai, Xiaoxuan Jia, Nicholas Turk-Browne,
  Kenneth Norman, James~J DiCarlo, and Daniel~LK Yamins.
\newblock How well do unsupervised learning algorithms model human real-time
  and life-long learning?
\newblock In {\em Thirty-sixth Conference on Neural Information Processing
  Systems Datasets and Benchmarks Track}.

\end{thebibliography}
}

\end{document}